\documentclass[10pt,twocolumn,letterpaper]{article}

\usepackage[pagenumbers]{cvpr} 
\usepackage{multirow}
\usepackage[T1]{fontenc}

\definecolor{cvprblue}{rgb}{0.21,0.49,0.74}
\usepackage[pagebackref,breaklinks,colorlinks,allcolors=cvprblue]{hyperref}

\def\paperID{9131} 
\def\confName{CVPR}
\def\confYear{2025}

\title{INFP: Audio-Driven Interactive Head Generation in Dyadic Conversations}

\author{
Yongming Zhu\thanks{Equal Contribution.} \quad Longhao Zhang\footnotemark[1] \quad Zhengkun Rong\footnotemark[1] \quad Tianshu Hu\footnotemark[1] \thanks{Corresponding author.} \quad Shuang Liang \quad Zhipeng Ge \\
Bytedance\\
{\tt\small \{zhuyongming, zhanglonghao.zlh, rongzhengkun, tianshu.hu, } \\
{\tt\small liangshuang.echo, zhipengge\}@bytedance.com}
}

\begin{document}

\twocolumn[{%
\renewcommand\twocolumn[1][]{#1}%
\maketitle
\vspace{-20pt}
\begin{center}
    \centering
    \captionsetup{type=figure}
    \includegraphics[width=0.99\textwidth]{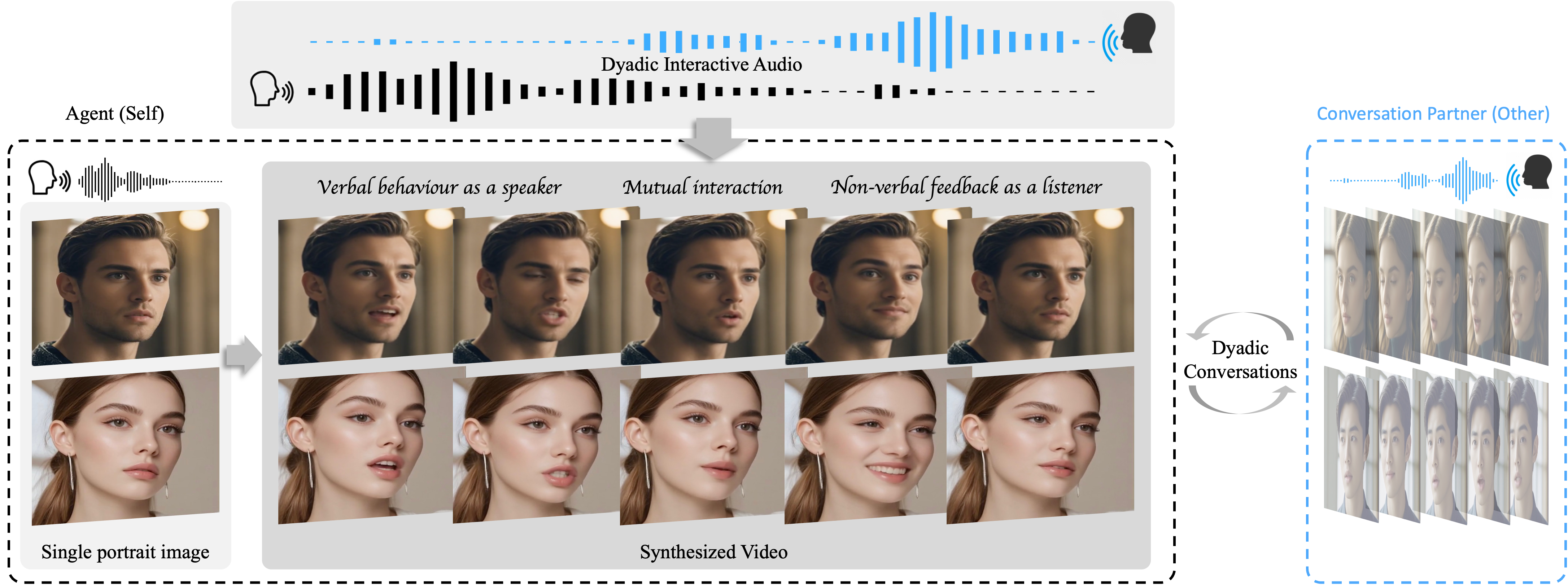}
    \captionof{figure}{We present INFP, an audio-driven interactive head generation framework for dyadic conversations. Given the dual-track audio in dyadic conversations and a single portrait image of arbitrary agent, our framework can dynamically synthesize verbal, non-verbal and interactive agent videos with lifelike facial expressions and rhythmic head pose movements. Additionally, our framework is lightweight yet powerful, making it practical in instant communication scenarios such as the video conferencing. INFP denotes our method is \textbf{I}nteractive, \textbf{N}atural, \textbf{F}lash and \textbf{P}erson-generic.}
\end{center}
}]
{
  \renewcommand{\thefootnote}%
    {\fnsymbol{footnote}}
  \footnotetext[1]{Equal contribution.}
  \footnotetext[2]{Corresponding authors.}
}

\begin{abstract}

Imagine having a conversation with a socially intelligent agent. It can attentively listen to your words and offer visual and linguistic feedback promptly. This seamless interaction allows for multiple rounds of conversation to flow smoothly and naturally. In pursuit of actualizing it, we propose INFP, a novel audio-driven head generation framework for dyadic interaction. Unlike previous head generation works that only focus on single-sided communication, or require manual role assignment and explicit role switching, our model drives the agent portrait dynamically alternates between speaking and listening state, guided by the input dyadic audio. Specifically, INFP comprises a Motion-Based Head Imitation stage and an Audio-Guided Motion Generation stage. 
The first stage learns to project facial communicative behaviors from real-life conversation videos into a low-dimensional motion latent space, and use the motion latent codes to animate a static image.
The second stage learns the mapping from the input dyadic audio to motion latent codes through denoising, leading to the audio-driven head generation in interactive scenarios.
To facilitate this line of research, we introduce DyConv, a large scale dataset of rich dyadic conversations collected from the Internet.
Extensive experiments and visualizations demonstrate superior performance and effectiveness of our method.
Project Page: \href{https://grisoon.github.io/INFP/}{\emph{https://grisoon.github.io/INFP/}}.

\end{abstract}    
\section{Introduction}
\label{sec:intro}

In recent years, researchers have dedicated considerable attention to audio-driven head generation \cite{PC-AVS,ng2022learning,ng2023can,song2023emotional,tian2024emo,xu2024vasa,liu2024customlistener,wang2023agentavatar,DIM,zhou2023interactive,guan2023stylesync}, in order to build the conversational agent. However, most studies only focus on single-sided communication, such as talking or listening, ignoring the dyadic properties in human-human interaction. 
Talking-head generation \cite{xu2024vasa,zhang2023sadtalker,tian2024emo,chen2024echomimic,xu2024hallo,zhang2023dream,liu2024anitalker,wang2024v,RAD-NeRF,PC-AVS,Wav2Lip,MakeItTalk,zhang2024personatalk} aims at synthesizing speaker's facial animations based on their reference images and a driving audio. Although these works can produce vibrant videos with accurate lip synchronization, they solely emphasize the speaker's role and overlook the listener's feedback. 
Listening-head generation \cite{liu2024customlistener,ng2022learning,ng2023can,song2023emotional,liu2023mfr,zhao2022semantic,zhou2022responsive,react2024}, on the other hand, intends to respond to speaker's behaviors. However, these works restrict the listener's response solely to non-verbal facial movements, which is clearly far from real-life interaction scenarios. 
Some recent studies \cite{DIM,zhou2023interactive,wang2023agentavatar} have started exploring head generation for dyadic interaction, but they require manual role assignment between \emph{Listener} and \emph{Speaker} and cannot achieve smooth and natural role switching.
We believe that for a conversational agent, it should not be predetermined a role, but rather be able to freely switch between listening and speaking states based on the current dialogue context.
When having a conversation with the agent, you may interrupt or respond to it at any moment. At that point, it naturally transitions from a speaker to a listener, and vice versa. 

In this paper, we propose a novel audio-driven head generation framework for dyadic interaction, namely INFP. Unlike previous works, we do not divide the model into \textit{Speaker Generator} and \textit{Listener Generator}, or apply explicit role switching. The driven individual dynamically alternates between the speaking and listening state, guided by the dyadic audio. 
Specifically, INFP comprises a Motion-Based Head Imitation stage and an Audio-Guided Motion Generation stage. 
In the first stage, our model learns to extract communicative behavior, including non-verbal listening cues and verbal speaking patterns, from massive real-life conversational videos and encode them to a motion latent space. 
The motion latent codes are then used to animate a static portrait image to authentically imitate the character in the video. A well-crafted motion latent space should exhibit a high degree of disentanglement, which means that head pose, facial expression, and emotion should be decoupled from appearance. To this end, we perform facial structure discretization and facial pixel masking to the input.
While in the second stage, the model is trained to map the dyadic audio to the motion latent space pretrained in stage 1 to obtain the corresponding motion latent codes. It consists of an interactive motion guider and a conditional diffusion transformer. 
The former takes as input the audio from both agent and its conversation partner, and retrieves verbal and non-verbal motions from learnable memory banks to construct the interactive motion feature.
The latter utilizes the interactive motion feature as the condition and, together with other signals, generates the motion latent codes through denoising.

\begin{figure}[htp]
    \centering


    \centering
    \includegraphics[width=0.48\textwidth]{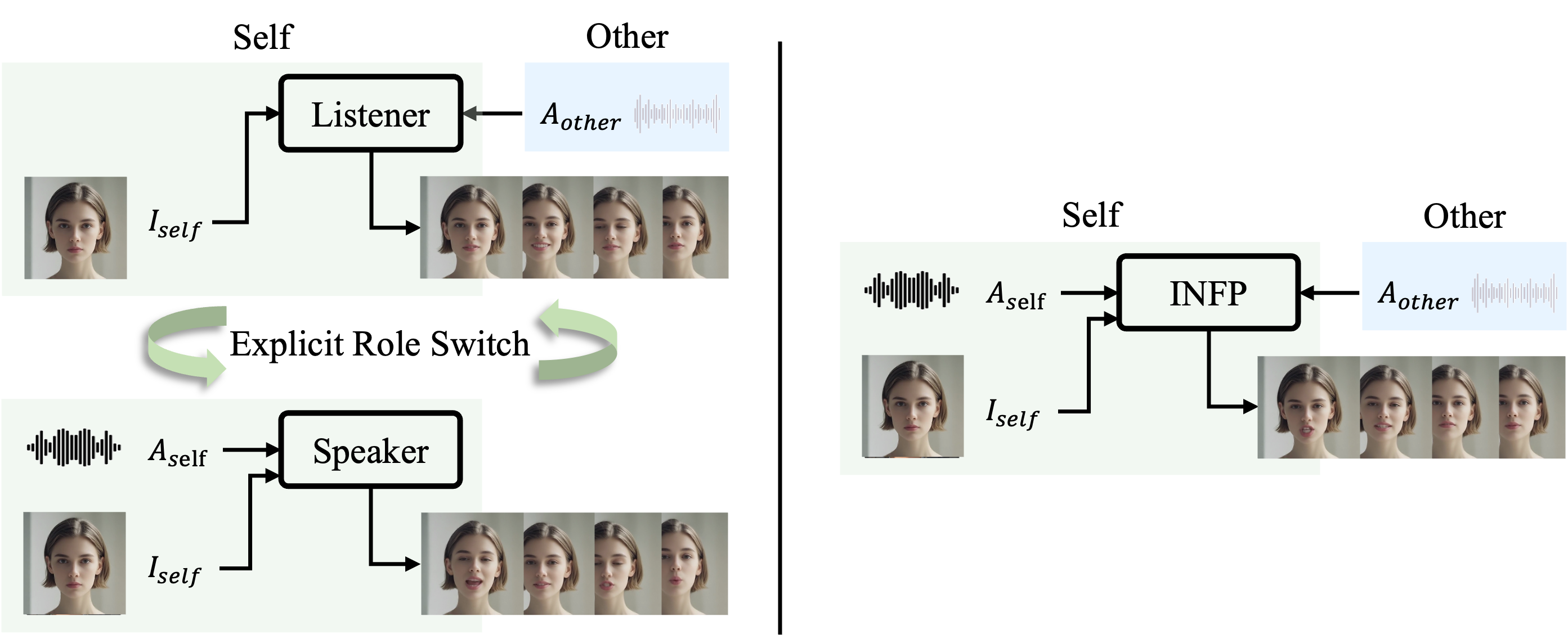}
    \label{fig:task}

    \caption{Objective illustration. Existing interactive head generation (left) applied manual role assignment and explicit role switching. Our proposed INFP (right) is a unified framework which can dynamically and naturally adapt to various conversational states.}
\end{figure}

To support our research on interactive head generation, we introduce DyConv, a large scale dataset of rich dyadic conversations collected from the Internet. It contains non-scripted video conversations between pairs of individuals from diverse backgrounds. The videos capture their discussions on a wide range of topics and emotions, portraying authentic communication in real-life settings. Compared to previous datasets like ViCo \cite{zhou2022responsive}, ViCo-X \cite{vicox} and RealTalk \cite{geng2023affective}, DyConv has a larger scale, higher quality, and richer interactions.

Our main contributions are summarized as follows:

\begin{itemize}
    \item We propose INFP, an audio-driven interactive head generation framework with a new paradigm. To our best knowledge, we are the first to enable the driven individual to naturally transition between listening and speaking states during multi-turn conversations, without the requirement for manual role assignment or explicit role switching.
    
    \item We design a novel interactive motion guider with learnable memory banks to adaptively extract interactive information from dual-track dyadic audio and construct speaking-listening mixed motions.
    
    \item We introduce a large-scale and high-quality dataset called DyConv. 
    We hope it will encourage the investigation of head generation for dyadic interaction.

    \item Our framework achieves real-time and person-generic head generation.
    Extensive experiments and visualizations demonstrate superior performance and effectiveness of our method. 
\end{itemize}

\section{Related Work}

\subsection{Single-Sided Audio-Driven Head Generation}

\textbf{Listening Head Generation.}
As a pioneer, RLHG \cite{zhou2022responsive} contributes the ViCo dataset as a common benchmark and builds a baseline method with a sequential decoder.
Following that, PCH \cite{PCH} introduces the enhanced renderer to improve the visual quality of generated videos.
Song et al. \cite{react2024} organizes the REACT challenge to encourage the community to investigate the face reaction in offline and online setting, separately.
L2L \cite{ng2022learning} leverages the VQ-VAE \cite{vqvae} to store head motion sequences in discrete codebooks, leading to the better motion diversity.
ELP \cite{song2023emotional} rearranges the latent space based on emotional priors to emphasize the emotion of generated motions.
To get rid of simple emotional labels, CustomListener \cite{liu2024customlistener} proposes to control listener head motions via the text prior.
MFR-Net \cite{liu2023mfr} makes a progress of identity preservation and motion diversity.
RealTalk \cite{geng2023affective} proposes a a medium-scale dataset and attempts to retrieve possible videos of listener head with large language model.
Ng et al. \cite{ng2023can} presents a transfer-based approach from pretrained large language models to predict the motion sequences.

\textbf{Talking Head Generation}
Audio-driven talking head generation has become a prominent research area in recent years. Early GAN-based approaches, like Wav2Lip \cite{Wav2Lip}, PC-AVS \cite{PC-AVS}, and MakeItTalk \cite{MakeItTalk}, directly integrate the input audio and video to produce lip-synced visuals. 
VASA-1 \cite{xu2024vasa} takes a significant leap forward in the field by decoupling latent feature spaces, enabling the generation of more realistic video outputs. This model can operate on a single image rather than requiring a full video sequence. 
With the emergence of diffusion models, several works \cite{DiffusedHead, tian2024emo, wang2024v} are proposed to enhance the generation quality.
EchoMimic \cite{chen2024echomimic} is concurrently trained using both audios and facial landmarks, and is capable of generating portrait videos not only by audios and facial landmarks individually, but also by a combination of both audios and selected facial landmarks.
AniTalker \cite{PC-AVS}, employing two self-supervised learning strategies, excels in capturing intricate facial movements. 

\subsection{Head Generation in Dyadic Conversations}
Recently, the research community has paid increasing attentions on the audio-driven head generation in the dyadic interaction.
As an extension of RHLG \cite{zhou2022responsive}, Zhou et al. \cite{zhou2023interactive} proposes a multi-turn conversational dataset, called ViCo-X, and designs a \emph{Role Switcher} to bridge the \emph{Listener Generator} and the \emph{Speaker Generator}. However, the explicit role switching could cause unnaturalness and inconsistency between different states. Furthermore, this kind of paradigm fails to cover all states in dyadic conversations, such as the agent and the conversation partner speak simultaneously.
DIM \cite{DIM} applies a pretraining approach that jointly models speakers’ and listeners’ motions to capture the dyadic context. In applications, the pretrained model needs to be additionally finetuned for downstream tasks like talking head and listening head generation, separately. To this end, the manual assignment of the role is necessary in dyadic conversations, leading to inappropriate transitions.
AgentAvatar \cite{wang2023agentavatar} designs a streamlined pipeline to predict behaviors of a photo-realistic avatar agent in dyadic interactions. While the motion synthesis is conditioned on the high-level textual description, AgentAvatar fails to generate facial behaviors that precisely align with the dialogue audio, as demo videos shown in their project page.
Additionally, there are some other studies in dyadic interactions \cite{sun2024beyond,ng2024audio,park2024let,huang2024interact,yan2023dialoguenerf}, but all of them are person-specific and suffer from the generalization ability to arbitrary individuals.
\section{Method}

\begin{figure*}
   \centering
   \includegraphics[width=0.99\linewidth]{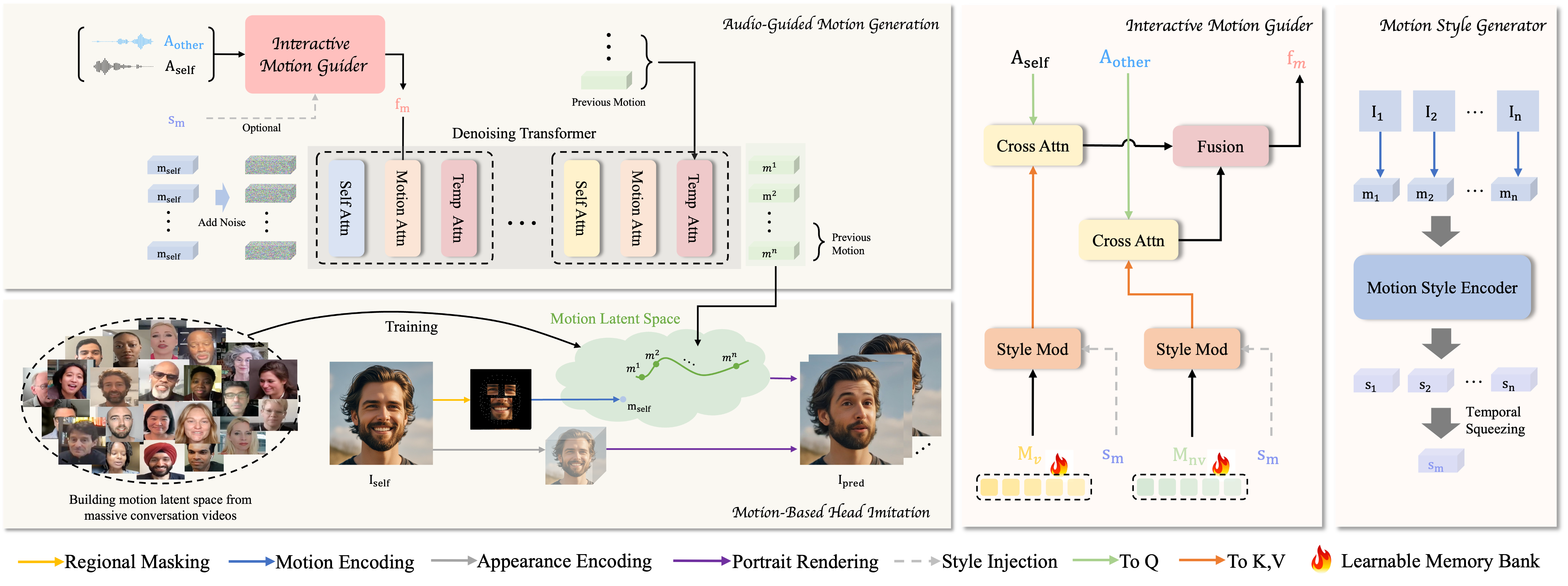}
   \caption{Overview of INFP. The first stage (Motion-Based Head Imitation) learns to project facial communicative behaviors from real-life conversation videos into a low-dimensional motion latent space, and use the latent codes to animate a static image. The second stage (Audio-Guided Motion Generation) learns the mapping from the input dyadic audio to motion latent codes through denoising, to achieve the audio-driven interactive head generation.}
   \label{fig:pipeline}
\end{figure*}

Our goal is to synthesize the agent head during a dyadic conversation. 
To this end, we propose a two-stage interactive head generation framework, called INFP. 

In the first stage, the model learns to extract and compress multiple conversational behaviors in videos into a low-dimensional motion latent space, and use these latent codes to animate the portrait image of the agent, denoted as $I_{self}$. Note that, to ensure the richness and diversity of the motion latent space, we train stage 1 with plenty of real-life dialogue videos.
Once the training converges, we freeze this part and start to train the second stage. The model takes as input the audio of the agent, denoted as $A_{self}$, and the audio of the conversation partner, denoted as $A_{other}$, to learn the mapping from this dual-track dialogue audio to motion latent codes through denoising. We present a comprehensive introduction to these two stages below.

\subsection{Motion-Based Head Imitation} 

\textbf{Motion Encoding.} For motion encoding, we employ a motion encoder $\textbf{E}_{m}$ to learn an implicit latent representation, which captures verbal and non-verbal communicative behaviors from input facial images. 
$\textbf{E}_{m}$ comprises a convolutional-based feature extractor followed by MLP layers, which compresses the motion feature into a latent code. 
Note that, we use implicit representation instead of explicit 3DMM coefficients \cite{ren2021pirenderer} because the expressiveness of 3DMM model is limited, and the commonly used expression coefficients, predicted by SOTA 3D facial estimation methods \cite{emoca,Deep3D}, entangle with face shape to a certain extent. 

A well-crafted motion latent space should demonstrate a high level of disentanglement to ensure that the extracted motions are appearance-independent. To achieve this, we apply two measures.
(a) We set the motion latent code as a low-dimensional 1-D descriptor. It fosters effectively capturing the fundamental semantics of facial motion without entangling with appearance information, as explained in previous works \cite{tishby2000information, wang2022latent, wang2023progressive}.
(b) We design a hybrid facial representation to replace the original input image of $\textbf{E}_{m}$.
In particular, we first mask the majority of facial pixels, preserving only the eyes and lip regions, as we regard them as the most intricate and expressive components of facial expressions. This can block the interference of irrelevant information such as hair and background for expression distill.
To provide facial orientation and contour information, we then use an off-the-shelf facial estimation model \cite{MediaPipe} to obtain the facial vertices, and project the contour-related points onto the masked image. Our ablation study shows the superiority of our proposed hybrid facial representation compared with the other representations like the intact image \cite{megaportrait} or 2D landmarks map \cite{ma2024followyouremoji}.

\textbf{Head Imitation.} In this paper, we adopt a motion flow-based generation pipeline. Concretely, we first extract source motion code $m_{self}$ from the input portrait image $I_{self}$, and driving motion codes $m_{dri}^{1:N}$ from the driving video $V_{dri}$.
A motion flow estimation model $\textbf{F}$ takes these motion codes as input to predict the ${self}\rightarrow{driving}$ flow.
We utilize a face encoder $\textbf{E}_{face}$ to extract 3D appearance feature volume from the input image $I_{self}$. This feature volume is then warped by the motion flow and sent into a face decoder $\textbf{D}_{face}$ to synthesize the final predictions $I_{pred}^{1:N}$. The overall pipeline can be written as:
\begin{equation}
    \begin{split}
         Flow_{{s}\rightarrow{d}} &= \textbf{F}(\textbf{E}_{m}(I_{self}), \textbf{E}_{m}(V_{dri})) \\
         I_{pred}^{1:N} &= \textbf{D}_{face}(\mathit{Warp}(\textbf{E}_{face}(I_{self}), Flow_{{s}\rightarrow{d}}))
    \end{split}
\end{equation}

\subsection{Audio-Guided Motion Generation}\label{sec:style}
Previous works \cite{zhou2023interactive,DIM,sun2024beyond} tend to divide their models into a \textit{Speaker Generator} and a \textit{Listener Generator}. Since the \textit{Speaker Generator} only receives speaking audio, while the \textit{Listener Generator} solely processes listening audio, it is necessary to segment the audio of both agent and its conversation partner into speaking clips and listening clips. Hence, in a multi-turn dialogue, explicit role switching is also crucial. 
We argue that it can lead to unnatural transitions in communicative states, particularly in real-life interaction contexts characterized by frequent role switching.
Besides, this design paradigm clearly goes against the intention of the conversational agent, as it should be able to naturally present different communicative states based on the progress of the conversation.

In this paper, we present a new paradigm, in which the model takes  as input the audio from both the agent and its conversation partner, eliminating the necessity for explicit role assignment or switching. The key innovations lie in extracting interactive information from dual-track dyadic audio to construct speaking-listening mixed motions, and mapping these motions to the motion latent space of stage 1 to achieve audio-driven head generation.

\textbf{Motion Feature Construction.} 
As shown in \cref{fig:pipeline}, an interactive motion guider is introduced to extract motion feature from the input audio. Within it, there are two memory banks $M_{v}$ and $M_{nv}$, storing verbal and nonverbal motion, respectively. Each motion memory bank consists of learnable embeddings to memorize typical motions, denoted as $e^{1:K}$, where
$e^{k}\in \mathbb{R}^{d}$ represents the $k$-th motion embedding and $d$ is the dimension. We set $K$ to 64 and $d$ to 512 in our experiments.
When agent is speaking, $A_{self}$ contains rich information, so we take it as $Query$ and obtain verbal motion feature from $M_{v}$, which is taken as $Key$ and $Value$, through a cross-attention layer. Since the partner is listening, $A_{other}$ carries very little information, hence no effective non-verbal motion feature can be obtained from $M_{nv}$. Therefore, after feature fusion, which involves element-wise sum and several MLP layers, verbal motion feature dominates and drives the agent to express the intended speaking state. However, when the partner starts speaking, $A_{self}$ becomes weak and $A_{other}$ fulfills a vital role. Accordingly, non-verbal motion feature from $M_{nv}$ dominates and forces the agent to present a listening state. In this way, our model can dynamically construct interactive motion feature $f_{m}$ based on the content of the dialogue audio, allowing the agent to convey the intended state.
Note that, $A_{self}$ and $A_{other}$ are encoded to the feature space through an off-the-shelf HuBERT audio encoder \cite{HuBERT}, before sent into the model.

Inspired by stylistic facial animation methods \cite{styletalk,sun2024diffposetalk}, we also introduce a motion style vector $s_{m}$ to explicitly edit the motion embeddings of memory banks, through a style modulation layer raised by StyleGAN2\cite{Karras2019stylegan2}. 
$s_{m}$ is first extracted from motion latent codes through a motion style encoder $\textbf{E}_{style}$, and then squeezed along the temporal dimension to a global descriptor.
We argue that $s_{m}$ contains some global information like emotions and attitudes, thus enhancing the authenticity and vividness of the generated results. 
Note that during training, $s_{m}$ is sourced from an arbitrary video clip of the driven individual, while in testing, $s_{m}$ can be extracted from any videos or set as null.

\textbf{Motion Mapping.}
To map the interactive motions $f_{m}$ to the pretrained motion latent space of stage 1, we employ a conditional diffusion transformer \textbf{T} \cite{ho2020denoising}. Given the data distribution $q(m^{1:N}, f_{m})$ where $m^{1:N}$ represents the corresponding motion latent codes with $N$ frames, the diffusion model approximates the conditional distribution $q(m^{1:N} | f_{m})$. $\textbf{T}$ only has 4 blocks, which makes our framework lightweight enough to achieve real-time interaction. Each block consists of a self-attention layer, a motion-attention layer, and a temporal-attention layer. $\textbf{T}$ predicts the noise added to the ground-truth motion latent codes for each denoising step. Diffusion timestep is converted into the sinusoidal embedding and then concatenated to the noised motion latent codes at the time dimension. In the motion attention layer, we take the latent feature output from preceding self-attention layer as the $Query$, while the interactive motion feature $f_{m}$ is utilized as the $Key$ and $Value$. Inspired by AnimateDiff \cite{guo2023animatediff}, we implement a temporal layer to ensure a smooth transition between adjacent windows. In particular, we incorporate the last 10 frames of the generated motion latent codes from the previous window as the condition of the current one. Note that in the inference time, we add noise to the self motion latent codes $m_{self}^{1:N}$, which are extracted from $I_{self}$ and replicated into $n$ copies.

\textbf{Training Strategy.}
In the training stage, we randomly set style vector to null with a probability of 0.3, and randomly drop interactive motion feature and previous motion latent codes with a probability of 0.5.
Also note that we adopt a warm-up strategy for training stage 2. Specifically, we only select single-sided conversation clips from our training set as ``simple case'' to train the model for several epochs, which helps with the initialization of two memory banks. After that, we randomly sample multi-turn clips from entire dataset for the remaining training process.

\section{Dataset}
Existing dyadic interaction datasets have limitations in both scale and quality. ViCo \cite{zhou2022responsive} is a small-scale dataset with 1.6 hours and 96 IDs, and it lacks multi-turn conversation scenarios. Although ViCo-X \cite{vicox} features rich multi-turn conversations, the total duration is only 0.4 hours. Besides, its background is monotonous, all of which are green screen. RealTalk \cite{geng2023affective} is a medium-scale dataset capturing dyadic conversations between pairs of individuals totaling to 115 hours of online public videos. However, it does not restrict the resolution of faces in the video, leading to low face resolution in many videos, thus impacting the quality of model generation. 

We propose DyConv, a large-scale dataset of multi-turn dyadic conversations collected from the Internet. The dataset contains video clips with a duration of over 200 hours, captures a wide gamut of emotions and expressions. We employ an off-the-shelf face detection model \cite{s3fd}, and keep only frames where both individuals of the conversation are fully visible and their facial resolution is greater than $400\times400$. In addition, we utilize a SOTA speech separation model \cite{zhao2023mossformer2} to separate the audio of two persons in the conversation, denoted as $A_{p1}$ and $A_{p2}$. Then we run an active speaker detection model \cite{beliaev2020talknet} to match each audio clip to the corresponding face in the original video. 
We will publicly release the dataset, along with separated audios and annotations. We hope it will encourage the investigation of head generation for dyadic interaction.

\section{Experiment}

\begin{figure*}[htp]
   \centering
   \includegraphics[width=0.93\linewidth]{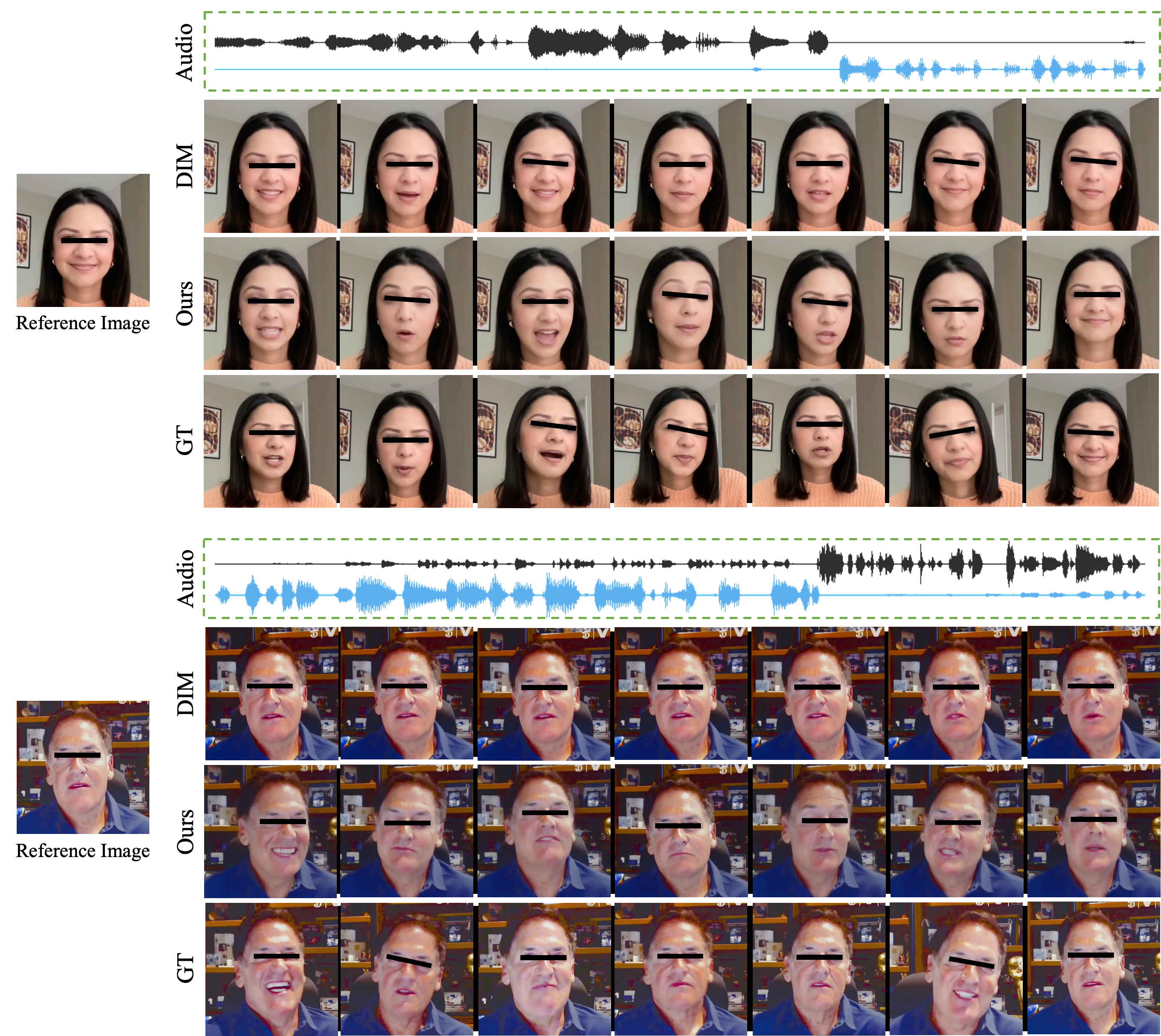}
   \caption{Qualitative comparison of interactive head generation on DyConv. Given the input audio of dyadic conversations (the audio of the predicted agent is black and the audio of the conversation partner is blue), our framework generates reasonable and vivid facial and head pose movements as various roles in interactive scenarios.}
   \label{fig:cmp-i}
\end{figure*}

\subsection{Implementation Details}
In our experiments, we train the diffusion transformer using AdamW optimizer \cite{loshchilov2019adamw}, with $\beta_{1}$ = 0.9 and $\beta_{2}$ = 0.999, a learning rate of 1e-4, a weight decay of 1e-2, and a batch size of 32.
For inference, we use a DDIM sampler \cite{song2020denoising} for 20 denoising steps. The CFG parameter of motion feature is set to 2, and the CFG parameter of previous motion latent is set to 1.5.


\subsection{Comparisons}

Although originally designed for dyadic interactive scenarios, our method can be naturally generalized to other applications, such as listening head generation and talking head generation.
This allows us to fully validate the advancement of our proposed framework against some of the current SOTA methods in related areas.

\textbf{Evaluation Metrics.} To quantitatively evaluate the visual quality, we utilize the Structured Similarity (SSIM), Peak Signal-to-Noise Ratio (PSNR) and Frechet Inception Distance (FID) for the generated videos. 
Additionally, LPIPS \cite{LPIPS} is employed to measure the feature-level similarity between generated and ground-truth frames.
To assess the preservation of facial appearance, we calculate the cosine similarity (CSIM) \cite{ArcFace} between the facial features of the reference image and the generated video frames.
In term of motion diversity of generated videos, we use SI for diversity (SID) and Variance (Var) following previous studies \cite{ng2022learning,DIM}.
We use the confidence score of SyncNet (SyncScore) \cite{SyncNet} to evaluate the synchronization between lip movements and audio signals of the agent.

\subsubsection{Interactive Head Generation}
Since only a few works focus on the interactive head generation \cite{wang2023agentavatar,zhou2023interactive} and these methods are not open-sourced, direct comparisons with them are considerably difficult. To effectively validate our method, we select the SOTA dyadic modeling work DIM \cite{DIM} as our baseline and utilize the officially released code to re-train the model on our datasets.
Specifically, we first use the whole dataset to pretrain the VQ-VAE for the listener and speaker and then split the dataset into two subset to train the listener generator and the speaker generator, respectively. To 
better visualize results of DIM, we also train the face renderer \cite{ren2021pirenderer} to generate photo-realistic videos.

\textbf{Quantitative Results.}
The quantitative experiments are carried out on the test set of DyConv. Please refer to \cref{table:cmp-i} for the results. It can be seen that our method outperforms the baseline cross a variety of metrics.
In particular, our face decoder achieves a better visual quality such as SSIM, PSNR and FID.
Thanks to the disentanglement and expressiveness of our latent space, our method is superior to DIM in term of audio-lip-synchronization by a large margin as there is a much closer distance of our SyncScore to the ground truth.
As for the LPIPS and CISM, our method still achieves solid performances, reflecting the better identity preservation.
Of particular note, our method is far ahead of the baseline in term of SID and Var, which proves the generalization and effectiveness of our method to generate a great range of diverse motions. It benefits from the design of interactive motion guider as well as conditional denoising transformer.

\begin{table*}[t]
    \centering
    \begin{tabular}{l c c c c c c c c c} 
        \toprule
        Methods &SSIM$\uparrow$ &PSNR$\uparrow$ &FID$\downarrow$ &SyncScore$\uparrow$ &LPIPS$\downarrow$ &CSIM$\uparrow$ &SID$\uparrow$ &Var$\uparrow$  \\ 
        \midrule
        DIM \cite{DIM}  &0.651 &20.417 &34.361 &4.778 &0.485 &0.824 &0.766 &0.825\\
        INFP (Ours)  &\textbf{0.834} &\textbf{31.562} &\textbf{15.727} &\textbf{7.188} &\textbf{0.257} &\textbf{0.904} &\textbf{2.613} &\textbf{2.386} \\
        \midrule
        w/o Motion Memory &0.830 &31.218 &18.334 &6.103 &0.259 &0.899 &2.153 &2.016 \\
        w/o Style Modulation &0.831 &31.442 &16.029 &7.062 &0.271 &0.904 &2.551 &2.316 \\
        w/ Intact Image &0.802 &28.488 &16.990 &6.812 &0.266 &0.842 &2.470 & 2.148 \\
        w/ Landmarks Map &0.821 &30.693 &16.327 &6.833 &0.281 &0.901 &2.601 & 2.335 \\
        \midrule
        GT  &1.000 &N/A &0.000 &7.261 &0.000 &0.967 &2.891 &2.435 \\
          \bottomrule
    \end{tabular}
    \caption{Quantitative comparison of interactive head generation on DyConv.}
    \label{table:cmp-i}
\end{table*}

\textbf{Qualitative Results.} Subjective evaluation is crucial in identifying the performance of generative models, particularly on videos. We strongly recommend readers to watch our supplementary video.
Two examples and their comparisons with the baseline are shown in \cref{fig:cmp-i}. We use two typical audio clips in the dyadic conversation scenario to drive reference identities randomly selected from the test set.
It can be obviously observed that our method can generate more diverse head pose movements in comparison to DIM.
Meanwhile, results of our method demonstrate more vivid facial expressions, which significantly improve the naturalness of generated videos. 
It is worth to noting that facial behaviors synthesized by our method are well aligned with the dialogue audio content. This suggests the effectiveness of our method in dyadic interactions, which is capable of capturing distinct characteristics of various states, such as listening, speaking as well as the mutual interaction, and naturally adapting its role among them based on the progress of the conversation.
In contrast, DIM is unable to generate appropriate behaviors such as reasonable lip movement and response feedback.

\textbf{User Study.} We also invite 20 participants to conduct a user study for further subjective evaluation.
Specifically, we apply INFP and DIM to synthesize results of 20 dyadic conversation videos randomly sampled from the test set.
By adopting the commonly used Mean Opinion Scores (MOS) rating protocol,
we request all the participants to provide their ratings (from
1 to 5, the higher the better) on four aspects for each generated video: naturalness, motion diversity, audio-visual alignment and visual quality. As detailed in \cref{table:user-study}, our method demonstrates superior performances across all evaluated aspects compared to DIM.

\begin{table}[t]
    \small
    \footnotesize
    \setlength{\tabcolsep}{4pt}
    \centering 
    \begin{tabular}{l c c c c c c c c c} 
        \toprule
        Methods  & Naturalness & Motion Diversity & AV-Align & Visual \\ 
            \midrule
            DIM \cite{DIM} & 2.71  & 2.14 & 2.65 & 3.13  \\
            INFP w/o Style & 4.02  & 3.98 & 4.28 & 4.11 \\
            INFP (Ours) & \textbf{4.38}  & \textbf{4.49} & \textbf{4.33} & \textbf{4.13} \\
          \bottomrule
    \end{tabular}
    \caption{User study measured by Mean Opinion Scores.}
    \label{table:user-study}
\end{table}

\subsubsection{Listening Head Generation}
Since our method can generalize to the task of listening head generation, we directly use our model without any modification to conduct additional experiments. Specifically, we compare our framework with SOTA listening head generation methods, including L2L \cite{ng2022learning}, RLHG \cite{zhou2022responsive} and DIM \cite{DIM}. ViCo \cite{zhou2022responsive} is selected as the benchmark.
For quantitative evaluation, we utilized the same test set and inherited both quantitative and qualitative results presented in DIM. We conducted the comparison on FD, MSE, SID and Var and each metric is calculated on expression and head pose, respectively. Different to other methods, our method generate photo-realistic video rather than 3DMM coefficients. To this end, we use the off-the-shelf 3D face reconstruction method \cite{emoca} to extract expression and head pose coefficients from our generated videos.
As shown in \cref{table:cmp-listen}, INFP outperforms other SOTA methods in the majority of metrics. While SID and Var are performing well, our method remains highly effective in both FD and MSE.
This indicates the power of our novel framework design and dataset quality, making generated facial expressions and head pose movements closer to real human behaviors. 
The visualization results in the supplementary materials also demonstrate that INFP achieves more expressive and diverse motions.

\begin{table}[t]
    \small
    \footnotesize
    \setlength{\tabcolsep}{4pt}
    \centering
    \begin{tabular}{l c c c c c c c c c} 
        \toprule
        \multirow{2}[2]{*}{Method}  & \multicolumn{2}{c}{FD$\downarrow$} & \multicolumn{2}{c}{MSE$\downarrow$} & \multicolumn{2}{c}{SID$\uparrow$} & \multicolumn{2}{c}{Var$\uparrow$} \\ 
        & Exp & Pose & Exp & Pose & Exp & Pose & Exp & Pose \\
        \midrule
        L2L \cite{ng2022learning} &33.93 &\textbf{0.06} &0.93 &0.01 &2.77 &2.66 &0.83 &0.02 \\
        RLHG \cite{zhou2022responsive} &39.02 &0.07 &0.86 &0.01 &3.62 &3.17 &1.52 &0.02 \\
        DIM \cite{DIM} &23.88 &\textbf{0.06} &0.70 &0.01 &3.71 &2.35 &1.53 &0.02 \\
        \midrule
        INFP (Ours) &\textbf{18.63} &0.07 &\textbf{0.51} &\textbf{0.01} &\textbf{4.78} &\textbf{3.92} &\textbf{2.83} &\textbf{0.18} \\
        \bottomrule
    \end{tabular}
    \caption{Quantitative comparison of listening head generation on ViCo \cite{zhou2022responsive}.}
    \label{table:cmp-listen}
\end{table}

\subsubsection{Talking Head Generation}
For talking head generation, we select SadTalker\cite{zhang2023sadtalker}, AniTalker\cite{liu2024anitalker}, and EchoMimic\cite{chen2024echomimic} as comparing methods.
As all of them reported evaluation results on HDTF \cite{HDTF}, we randomly sampled 20 videos from it as the test set to conduct our experiments for a fair comparison.
Results in \cref{table:cmp-talk} show that our model is superior in audio-lip-synchronization and identity preservation.
The reason behind this might be that we carefully build a disentangled and expressive motion latent space and it makes our model easier to learn the correlation between audio and lip movements without the interference of appearance features. It can also be observed from the visualization results in the supplementary materials.
Our SSIM and PSNR score is slightly behind EchoMimic because EchoMimic is finetuned based on the Stable Diffusion V1.5 \cite{rombach2021highresolution}. Despite this, our method can achieve better scores in terms of FID, meaning a good visual quality of our results.

\begin{table}[t]
    \small
    \footnotesize
    \setlength{\tabcolsep}{4pt}
    \centering
    \begin{tabular}{l c c c c c c c c c} 
        \toprule
        Methods  &SSIM$\uparrow$ &PSNR$\uparrow$ & FID$\downarrow$  & SyncScore$\uparrow$  & CSIM$\uparrow$ \\ 
            \midrule
            SadTalker \cite{zhang2023sadtalker}  &0.781 &26.650 &28.516 &6.125 &0.814 \\
            AniTalker \cite{liu2024anitalker}  &0.789 &26.822 &28.778 &5.738 &0.858 \\
            EchoMimic \cite{chen2024echomimic} &\textbf{0.807} &\textbf{28.983} &20.873 &6.251 &0.863 \\
            \midrule
            INFP (Ours) &\textbf{0.804} &28.657 &\textbf{18.582} &\textbf{6.670} &\textbf{0.886} \\
          \bottomrule
    \end{tabular}
    \caption{Quantitative comparison of talking head generation on HDTF \cite{HDTF}.}
    \label{table:cmp-talk}
\end{table}


\subsection{Ablation Study}
We ablate 3 key components of our framework, and the results are listed in \cref{table:cmp-i}. 

\textbf{Motion Memory Bank.}
We remove the verbal and non-verbal motion memory banks from interactive motion guider in stage 2, and directly use the dyadic audio feature after several MLP layers as the condition for the motion-attention layer. Results show that the generated lip-related motion behaviors are clearly inconsistent with the ground-truth.
In additional, there is a significant drop of both SID and Var scores. These suggest that memory banks can effectively store rich motion patterns in different conversational states.

\textbf{Style Modulation.}
We remove the motion style vector $s_{m}$ and corresponding modulation layers. As we expected, the authenticity and vividness of the results decrease, which can also be reflected in the user study.

\textbf{Hybrid Facial Representation.} We change the input to $\textbf{E}_{m}$ in stage 1 from our carefully-designed hybrid facial representation to the original intact image or 2D landmarks map. We observe that there is degradation in generation quality and a leakage of appearance information.

\subsection{Visualization of Style Control}
\begin{figure}[htp]
   \centering
   \includegraphics[width=0.9\linewidth]{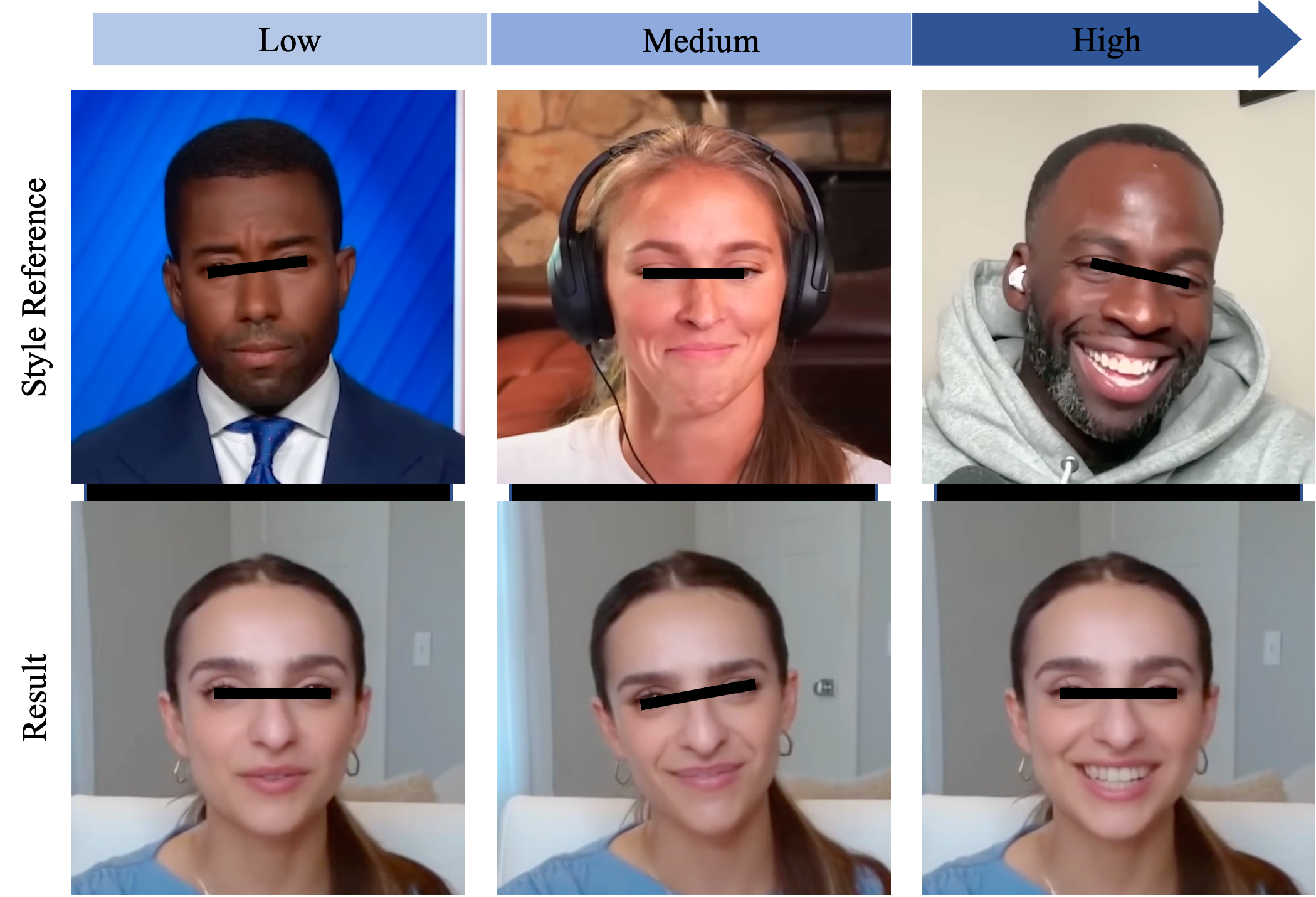}
   \caption{Visualization of style control with our style modulation.}
   \label{fig:style}
\end{figure}

As illustrated in \cref{sec:style}, our model supports inputting style vector $s_{m}$ extracted from any portrait video to globally control the emotions or attitudes of the generated results. Here we select 3 video clips with varying degrees of ``happiness'' to control emotions during generation. Results are shown in \cref{fig:style}. It can be seen from the figure that $s_{m}$ effectively control the emotions of the driven individual.

\section{Conclusion}
In this paper, we propose INFP, a novel audio-driven interactive head generation framework for dyadic conversations, which aims to mimic facial behaviors of human-human interactions conditioning on the dyadic audio.
As demonstrated, our method is a unified framework and is capable of smoothly and naturally adapting to various conversational roles without explicit assignment or switching.
Meanwhile, we collect a large scale audio-visual dataset for multi-turn dyadic conversations, which could facilitate further researches of dyadic interactive head generation.
Extensive experiments and ablations prove the advancement and effectiveness of our method over other SOTA methods.

\textbf{Limitation and Future Work.} Currently, our method only takes conversation audios as the input, combining control signals from multiple modalities, such as visual and textual contents, can offer additional capacities. As our proposed framework focus on head synthesis, extending the generation range to the upper, even whole body might be a interesting and challenging future work.

\textbf{Ethical Considerations.} There might be the potential misuse of our method to fabricate deceptive talks and speeches. To this end, we will restrict access to our core models exclusively to research institutions.

\section*{Acknowledgment}
We would like to thank Lizhen Wang for his review
of this article and his constructive feedback.

{
    \small
    \bibliographystyle{ieeenat_fullname}
    \bibliography{main}
}


\end{document}